\documentclass[letterpaper, 10 pt, conference]{ieeeconf}  

\usepackage{paralist} 
\usepackage{booktabs} 
\usepackage{color} 
\usepackage{amssymb}
\usepackage{graphicx}
\usepackage{multirow}
\usepackage{amssymb}
\usepackage{graphicx}
\usepackage{amsmath}
\usepackage{algorithm}
\usepackage{algpseudocode}
\usepackage{url}
\renewcommand{\vec}[1]{\boldsymbol{#1}}

\newcommand{\mat}[1]{\boldsymbol{#1}}

\newcommand\newnotecommand[3]{%
\newcommand#1[1]{{\color{#3}\footnote{{\color{#3}#2:} ##1}}}}
\newnotecommand\joni{Joni}{red}
\newnotecommand\anders{Anders}{cyan}
\newnotecommand\antti{Antti}{blue}

\IEEEoverridecommandlockouts                              

\title{\LARGE \bf
  AR-based interaction for safe human-robot collaborative manufacturing
}

\author{Antti Hietanen$^{1}$, Jyrki Latokartano$^{2}$, Roel Pieters$^{3}$, Minna Lanz$^{2}$ and Joni-Kristian K\"am\"ar\"ainen$^{1}$%
\thanks{Laboratories of Signal Processing$^{1}$, Mechanical Engineering and Industrial Systems$^2$ and Automation and Hydraulic Engineering$^3$, Tampere University of Technology, Finland {\tt\small First.Family@tut.fi}}%
}

\begin{document}

\maketitle
\thispagestyle{empty}
\pagestyle{empty}

\begin{abstract}
  Industrial standards define safety requirements for
  Human-Robot Collaboration (HRC) in industrial manufacturing. The standards
  particularly require real-time monitoring and securing of
  the minimum protective distance between a robot and an operator. In this work,
  we propose a depth-sensor based model for workspace monitoring and an
  interactive Augmented Reality (AR) User Interface (UI) for safe HRC. The AR UI
  is implemented on two different hardware: a projector-mirror setup and
  a wearable AR gear (HoloLens). We experiment the workspace model and
  UIs for a realistic diesel motor assembly
  task. The AR-based interactive UIs provide 21-24\% and 57-64\% reduction
  in the task completion and robot idle time, respectively, as compared to
  a baseline without interaction and workspace sharing.
  However, subjective evaluations reveal
  that HoloLens based AR is not yet suitable for industrial manufacturing while
  the projector-mirror setup shows clear improvements in safety
  and work ergonomics.
\end{abstract}

\section{INTRODUCTION}
\noindent
Industrial manufacturing is going through a process of change toward flexible and intelligent manufacturing, the so-called Industry 4.0. 
Human-robot collaboration (HRC) will have a more prevalent role and this evolution means breaking with the established safety procedures as the separation of workspaces between robot and human operator is removed.
However, this will require special care for human safety as the existing industrial standards and practices are based on the principle that operator and robot workspaces are separated and violations between them are monitored.

The International Organization for Standardization
(ISO) Technical Specification (TS) 15066~\cite{isots15066} has recently addresses this need
for safety with industrial collaborative robotics and defines {\em four different
collaborative scenarios}. The first specifies the need and required performance for
a safety-rated, monitored stop (robot moving is prevented without an emergency
stop conforming to the standard). The second outlines the behaviors expected for
hand-guiding a robot's motions via an analog button cell attached to the robot.
The third specifies the minimum protective distance between a robot and an
operator in the collaborative workspace, below which a safety-rated, controlled
stop is issued. The fourth limits the momentum of a robot such that contact with
an operator will not result pain or injury. Our work focuses on the third scenario
where the operator-robot distance is communicated interactively.

\begin{figure}[t]
  \begin{center}
    \includegraphics[width=0.8\linewidth]{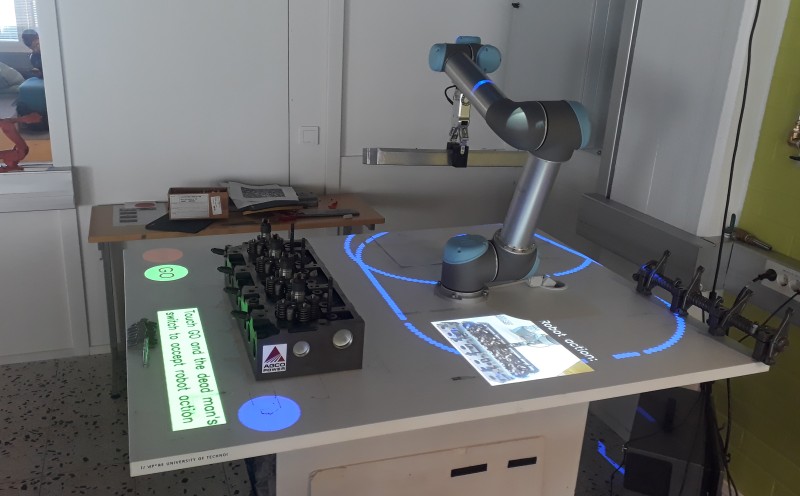}
    
    \vspace{0.5\medskipamount}
    
    \includegraphics[width=0.8\linewidth]{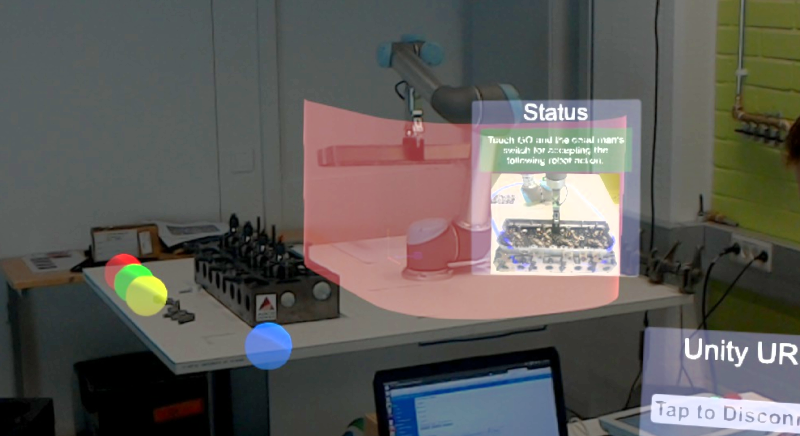}
    \caption{The proposed interactive UIs for safe human-robot manufacturing: projector-mirror (top) and HoloLens (bottom).
      \textcolor{blue}{Video: https://youtu.be/A8VTTcYFoMo}}
    \label{fig:pull_figure}
  \end{center}
\end{figure}

Marvel and Norcross~\cite{Marvel-2017-rcim} propose methods for the ISO/TS 15066
defined human-robot speed and separation monitoring based on measurable
quantities (robot and operator speeds and uncertainties).
The main focus of
this work is to define a model to monitor safety margins with a depth sensor and
to communicate the margins to the operator with an interactive User
Interface (UI) (Figure~\ref{fig:pull_figure}).

We propose a shared workspace model for HRC
manufacturing and interactive UIs.
The model is based on the virtual zones introduced by
Bdiwi et al.~\cite{bdiwi2017new}: {\em robot zone} and {\em human zone}.
In the human zone
an operator can freely move and the robot is not allowed to enter.
The robot zone is dynamically changing based on robot tasks and if
the operator or any other object enters the robot zone, the robot
is halted. In the proposed model, the two zones are separated by
a safety monitored {\em danger zone} and any changes in the workspace
model, either from the robot or operator side, cause halting the robot.
The purpose of the safety zone is to allow dynamic update of
the workspace model 
without compromising safety.

\section{RELATED WORK}

\paragraph{Human-Robot Collaboration (HRC) in industrial manufacturing}
Manufacturing industry leans on industrial standards that define
safety requirements for Human-Robot Collaboration and therefore it is
important to reflect research to the existing standards. Hitherto the
main safety principle has been separation of human and robot workspaces
and monitoring their violations. However, Industry 4.0 requires
more flexible HRC and therefore the recent ISO/TS 15066~\cite{isots15066}
provides best practices for shared workspaces. Marvel and
Norcross~\cite{Marvel-2017-rcim} proposed methods for the ISO/TS 15066
compliant human-robot speed and separation monitoring. However,
shared workspaces are yet only rarely used in industry as shown
in a recent survey~\cite{villani2018survey}.

There have been attempts to formalize HRC in industrial settings such
as Bdiwi et al.~\cite{bdiwi2017new} who introduce four levels of
HRC. The proposed workspace model, safety monitoring and User
Interfaces (UIs) in our work are consistent with their collaboration
levels Level~1 and Level~2. In Level~1 the robot and operator work close
to each other, but have their own tasks. In Level~2 there is an additional 
{\em cooperation zone} where the robot and operator collaborate without physical
interaction, for example, the robot holds a component firmly and stationary.
Level~3 defines a {\em handing-over zone} and Level~4 adds interaction where,
for example, human guides a heavy component held by the robot. From the
safety perspective, Lasota et al.~\cite{Lasota-2017} provide a survey of
existing HRC safety approaches and divide them into four categories:
Safety Through Control, Safety Through Motion Planning, Safety Through
Prediction and Safety Through Consideration of Psychological
Factors. Our works belongs to the Safety Through Control
category.

Safety through control is the most active research field in HRC
safety. There are existing works to stop a robot for
collision avoidance~\cite{lasota2014toward,flacco2015depth,vogel2017safeguarding} and motion re-planning~\cite{icinco15,unhelkar2018human}.
The proposed workspace model and safety monitoring are inspired
by the safety zones in~\cite{bdiwi2017new} and instead of a passive
system we propose AR-based interaction for HRC.

\vspace{\medskipamount}

\paragraph{Augmented Reality in HRC}
Shared workspace HRC requires seamless and safe interaction
between a robot and an operator. These have been
studied in several recent
works that augment the workspace with interactive and dynamic
user interface features~\cite{chadalavada2015s,andersen2016projecting,vogel2017safeguarding}.

Chadalavada et. al~\cite{chadalavada2015s} proposed a projector based safety
system for mobile robots operating on a flat industrial floor. The planned
navigation path was projected on the floor to increase human awareness.
Their user study verified that the projector system increased
predictability and transparency experience of co-workers in the same space
with the robots.

Andersen et al.~\cite{andersen2016projecting} proposed a projector based display
for HRC in industrial car door assembly.
User studies of the systems
against two baselines, a monitor display and simple text
descriptions, showed clear improvements in the terms of
effectiveness and user satisfaction.

Vogel et al.~\cite{vogel2011towards,vogel2013projection,vogel2017safeguarding}
have proposed multiple projector-camera based systems for safe HRC in a similar
setting to ours. \cite{vogel2011towards}
introduces a safety monitoring system based on a projector and multiple
camera based monitoring. The robot working area is projected by
a standard DLP projector and the violations are detected by geometric
distortions of the projected line due to depth changes.
In the more recent work~\cite{vogel2017safeguarding} the system is installed to an
industrial workcell with a shared screwing task. Our model is
based on a depth camera which make workspace model update and monitoring
more straightforward.

Wearable AR gear, also referred to as ``Augmented Reality Head-mounted Displays'',
have recently gained momentum.
Huy et al.~\cite{7942791} use a
head-mounted display in an outdoor application where a projector system cannot
be used. Elsdon et al.~\cite{Elsdon2018icra} introduced a handheld spray robot
where the control of the spraying was shared between the human and robot. However,
it is unclear how mature the wearable AR gear technology is for real industrial
manufacturing and therefore we compara wearable AR and projector-based AR
in our work.

\vspace{\medskipamount}

\noindent The main contributions of our work are:
\begin{compactitem}
  \item We propose a depth-based model for shared workspace Human Robot Collaboration. The model is based on three zones, human, robot and danger, and their continuous update and safety monitoring.
  \item We propose two AR-based and interactive User Interfaces (UIs) based on the proposed model: projector-mirror and wearable AR gear (HoloLens).
\end{compactitem}
We experimentally evaluate the model and UIs on a realistic industrial assembly task and report results from quantitative and qualitative (subjective)
evaluations with respect to performance, safety and ergonomy, and against
a non-shared workspace baseline.

\section{The shared workspace model}
\label{sec:safety_model}

\begin{figure}[h]
  \begin{center}
    \includegraphics[width=0.6\linewidth]{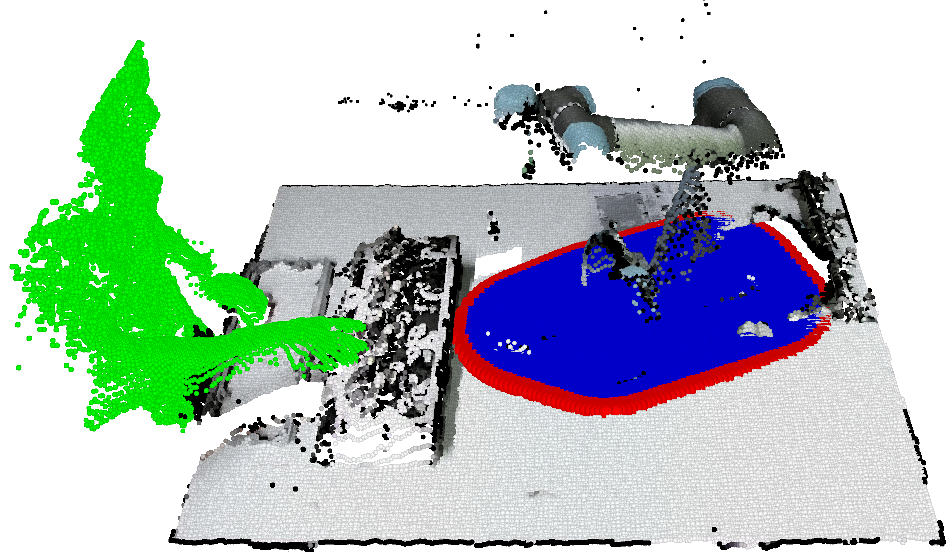} \\
    \includegraphics[width=0.4\linewidth]{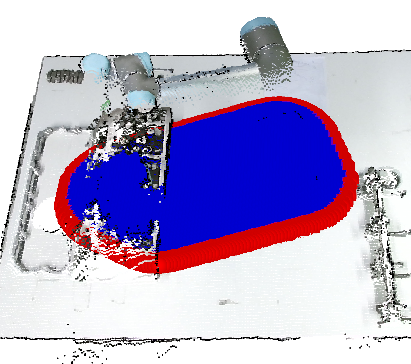} 
    \includegraphics[width=0.4\linewidth]{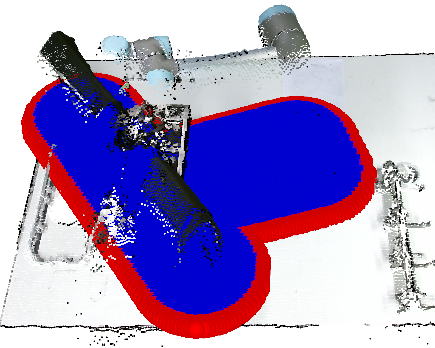}
    \caption{A shared workspace $S$ is modelled as a depth
      map image $I_S$ aligned with the robot coordinate system.
      The robot zone $Z_r$ (blue) is dynamically updated and subtracted
      from $I_S$ to generate the human zone $Z_h$ (gray). The two zones
      are separated by the danger zone $Z_r$ (red) which is monitored for safety violations.
      Changes in $Z_h$ are recorded to binary masks $R_i$ (green).
      Manipulated objects are automatically added to $Z_r$ (bottom right).}
    \label{fig:zones}
  \end{center}
\end{figure}
In our model, a shared workspace $S$ is modelled with a single depth map image $I_S$
and divided to three virtual zones:
robot zone $Z_r$, human zone $Z_h$ and danger zone $Z_d$ (Figure~\ref{fig:zones}). The zones
are modelled by binary masks in the same space as $I_S$ which makes
their update, display and monitoring fast and simple.

\subsection{Depth-based workspace model}

We consider a shared workspace monitored by a depth sensor
 which can be modelled as a pin-hole camera  parametrized by two matrices: the intrinsic camera matrix $\mat{K}$, modelling the projection of a Cartesian point to a image plane, and the extrinsic camera matrix $(\mat{R}~|~\mat{t})$, describing the pose of the camera in the world. The matrices can be solved by the chessboard calibration procedure \cite{park_robot_1994}. For simplicity we use the robot coordinate frame as the world frame.

After calibration
the points $\vec{p}$ in the depth sensor plane can be transformed to a Cartesian point in world frame and finally to the workspace model $I_S = \{\vec{x}\}_i$ of the size $W \times H$:
\begin{equation}
  \vec{P} = \mathcal{N}^{-1}\left( \mat{R}\mat{K}^{-1}
  \vec{p}
  +\vec{t}\right)
  \label{eq:inversemapping}, 
\end{equation}
\begin{equation}
\vec{x} = \mathcal{T}_{proj}\mat{P},
\label{eq:mapping}
\end{equation}
where $\mathcal{N}^{-1}$ is the inverse coordinate transformation and $T_{proj}$ is the projective transformation. Now, computations are done
efficiently in $I_S$ and (\ref{eq:inversemapping}) is used to display
the results to the AR hardware and (\ref{eq:mapping}) to map the
robot control points (Sec. \ref{sec:zones})  to the workspace model.

\subsection{Binary zone masks}
\label{sec:zones}
Since all computation is done in the depth image space
$I_S$ the three virtual zones can be defined as binary
masks of the size $W \times H$: the robot zone
$Z_r$, the danger zone $Z_d$ and the human zone $Z_h$.

\paragraph{The robot zone mask $Z_r$}

The zone is initialized using set of control points $C_r$ containing minimum number of 3D points covering all the extreme parts of the robot. 
The point locations in the robot frame are calculated online using a modified version of the Hawkinses model \cite{hawkins2013analytic} and projected to $I_S$.  %
Finally, the projected points are converted to regions having radius of $\omega$ and a convex hull \cite{graham1983finding} enclosing all the regions is computed and the resulting hull is rendered as a binary mask $M_r$ representing $Z_r$.

\paragraph{The danger zone mask $Z_d$}
Contour of the $Z_r$ and constructed by
adding a danger margin $\Delta\omega$ to the robot zone mask
and then subtracting $Z_r$ from the results:
\begin{equation}
Z_d = M_r(\omega+\Delta\omega) \setminus Z_r.
\end{equation}

\paragraph{The human zone mask $Z_h$} This is easy to
compute as a binary operation since the human zone is all
pixels not occupied by the robot zone $Z_r$ or the
danger zone $Z_d$:
\begin{equation}
Z_h = I_S \setminus  (Z_r \cup Z_d).
\end{equation}
\subsection{Adding the manipulated object to $Z_r$ and $Z_d$}
An important extension of our model is that the known objects
that the robot manipulates are added to the robot zone
$Z_r$ and $Z_d$ (see Figure~\ref{fig:zones}). This guarantees that the
robot does not accidentally hit the operator with an object
it is carrying.
In such case a new set of control points $C_{obj}$ is created using known dimensions of the object and point locations of the robot joints.
Finally the binary mask $M_{obj}$ for the object is created similarly as $M_r$ and the final shape of the zones are computed by fast binary operations:

\begin{equation}
Z_r = M_r(\omega) \cup M_{obj}(\omega), 
\end{equation} 
\begin{equation}
Z_d = M_r(\omega+\Delta\omega) \cup M_{obj}(\omega+\Delta\omega) \setminus Z_r \enspace .
\end{equation}

\subsection{Safety monitoring}
The main safety principle is that the depth values in
the danger region $Z_d$ must match with the stored depth
model. Any change must produce immediate halt of the system.
Our depth based model in the robot frame $I_S$ provides now
fast computation since the change detection is computed
as a fast subtraction operation
\begin{equation}
  I_{\Delta} = ||I_S - I||.
\end{equation}
where $I$ is the most recent depth data transferred to same space as our workspace model. The difference bins (pixels) are  further processed
by Euclidean clustering \cite{rusu2010semantic} to remove spurious bins due to noisy
sensor measurements. 

Finally, the safety operation depends on which zone a change
is detected:
\begin{equation}
\forall \vec{x} ~\mid~ I_{\Delta}(\vec{x}) \ge \tau
\begin{cases}
\hbox{ if } \vec{x} \in Z_d &\hbox{\em HALT}\\ 
\hbox{ if } \vec{x} \in Z_r &I_S(\vec{x}) = I(\vec{x}) \\
\hbox{ if } \vec{x} \in Z_h
& \scalebox{0.87}[1]{$M_h = \mat{0},~M_h(\vec{x}) = 1$}
\end{cases} \enspace ,
\label{eq:zones}
\end{equation}
where $\tau$ is the depth threshold.

In the first case, the change has occurred in the danger
zone $Z_d$ and therefore the robot must be immediately halted
to avoid collision. For maximum safety this processing stage
must be executed first and must test all pixels $\vec{x}$ before the
next stages.

In the second case, the change has occurred
in the robot working zone $Z_r$ and is therefore caused by
the robot itself by moving and/or manipulating objects and
therefore the workspace model $I_S$ can be safely updated.

In
the last case, the change has occurred in the human safety
zone $Z_h$ and we create the mask $M_h$ that represents the
changed bins (note that the mask is recreated for every measurement to allow temporal changes, but it does not affect
robot operation). Robot can continue operation normally, but
if its danger zone intersects with any 1-bin in $M_h$, then these
locations must be verified from the human co-worker via user interface. If the bins
are verified, then these values are updated to the workspace
model $I_S$ and operation continues normally. Note that our
system does not verify each bin separately, but a spatially
connected region of changed bins. This operation allows a
shared workspace and arbitrary changes in the workspace
which do occur away from the danger zone.

\section{The user interfaces}
\label{sec:hrc_framwork}

\begin{figure}[h]
  \begin{center}
    \includegraphics[width=0.41\linewidth]{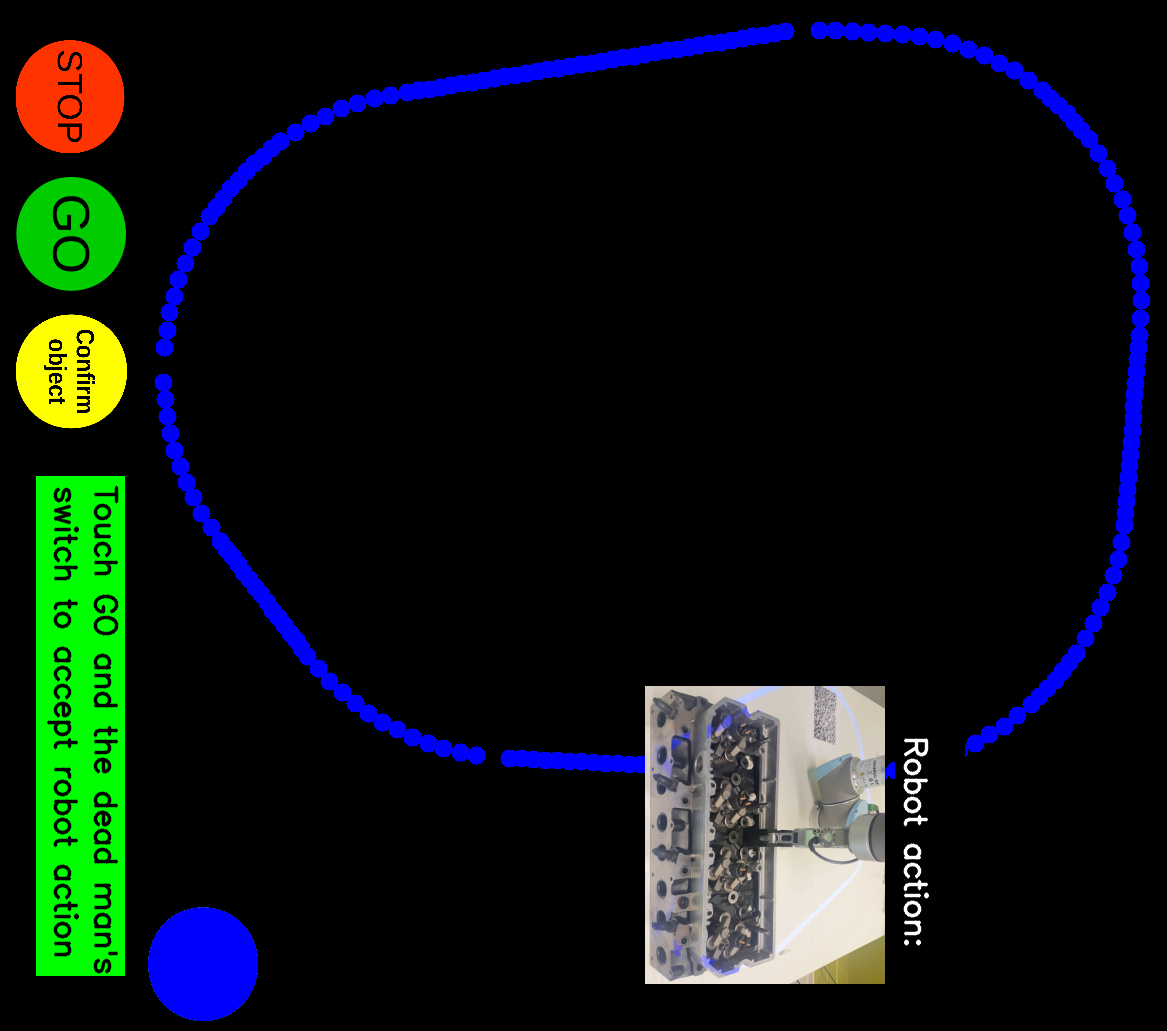}
    \includegraphics[width=0.55\linewidth]{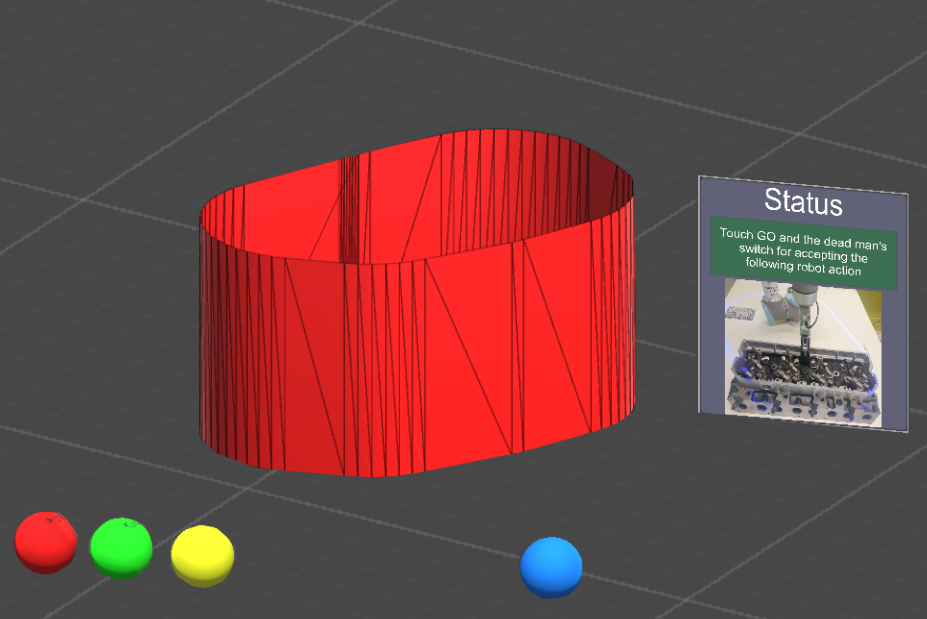}
    \caption{User Interface graphics for the projector-mirror as a 2D color image (left) and the HoloLens setup (right) rendered in Unity3D engine.}
    \label{fig:interface_view}
  \end{center}
\end{figure}

The danger zone defined in Section~\ref{sec:zones} and various User Interface (UI)
components are rendered to graphical objects in
two AR setups (Figure~\ref{fig:interface_view}).

\subsection{UI Components}
The proposed UI contains the following interaction components (Figure~\ref{fig:interface_view}
and Figure~\ref{fig:zones}):
1) a danger zone that shows the region operators should avoid;
2) highlighting changed regions in the human zone;
3) ``GO'' and ``STOP' buttons to start and stop the robot;
4) ``CONFIRM'' button to verify and add changed regions to the current model;
5) ``ENABLE'' button that needs to be pressed simultaneously with the ``GO'' and ``CONFIRM''
buttons to take effect; and
6) a graphical display box (image and text) to show the robot status and
instructions to the operator.

The above UI components were implemented to two different hardware, projector-mirror and
HoloLens. The UI components and layout were the same for the both hardware
to be able to compare the human experience on two different types of hardware.

\subsection{Projector-mirror AR}

The projector-mirror setup is adopted from Vogel et
al.~\cite{vogel2011towards,vogel2013projection,vogel2017safeguarding} with the main
difference that we replace multiple RGB cameras with a single RGB sensor (KinectV2).
A standard 3LCD projector is installed to the ceiling to point to a $45^\circ$ tilted mirror that
reprojects the picture to the workspace area. The mirror is needed to expand the projection
area of the standard projector, but could be replaced with a wide angle
lens projector. The projector outputs a $1920 \times 1080$ color image with 50Hz frame rate.
The projector coordinate frame is calibrated to the world (robot) coordinate frame
using the inverse camera calibration with a checkerboard pattern~\cite{MarKamLen:2011}.

\subsection{Wearable AR (HoloLens)}
As a state-of-the-art head-mounted AR display, we adopt Microsoft HoloLens.
The headset can operate without any external cables and the 3D reconstruction
of the environment as well as accurate 6-DoF localization of the head pose
is provided by the system utilizing an internal IMU sensor, four
spatial-mapping cameras, and a depth camera. The data exchange between HoloLens
and our model is done using wireless TCP/IP. We implemented a Linux server that
synchronizes data from the robot simulator (ROS) to HoloLens and back.

The interaction buttons are displayed as semi-transparent spheres that are positioned
similar to the projector-mirror UI (Figure~\ref{fig:pull_figure}). In addition,
we render the safety region as a solid virtual fence. The fence is rendered as
a polygonal mesh having semi-transparent red texture. From the 2D boundary and
a fixed fence height we construct the fence mesh from rectangular quadrilaterals that are
further divided to two triangles for the HoloLens rendering software.

The UI component and the virtual fence coordinates $\vec{p}$ are defined in the robot frame and
transformed to the HoloLens frame by
\begin{equation}
\vec{P'} = \mat{T}^{AR}_{H} \mat{T}^R_{AR} \vec{P}
\end{equation}
where $T^M_{R}$ is a known static transformation between the robot and an AR marker (set manually to the workspace) and $T^{AR}_{H}$ is the transformation
between the marker and the user holographic frame . Once the pose has been
initialized the marker can be removed and during run time $T^{AR}_{H}$ is updated by
HoloLens software.

\section{Engine assembly task}
\label{sec:task}
Our task is adopted from a local diesel engine manufacturing company and we
also present a baseline without a shared workspace.

\subsection{Task description}
\label{sec:task_desc}
\begin{figure}[h]
  \includegraphics[width=0.8\linewidth]{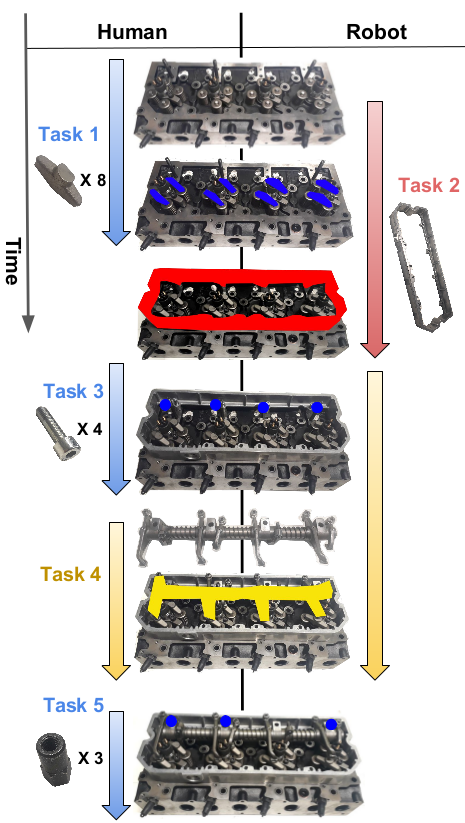}
  \caption{The engine assembly task used in the experiments.
    The task consists of five sub-tasks (Task~1-5) that
    are conducted by the operator (blue) or the robot (red) or both (yellow).
    Task 4 is the collaborative sub-task where
    a rocker shaft is held by the robot and carefully positioned by the
    operator.}
  \label{fig:assembly_steps}
\end{figure}

The task used in our experiments is a part of a real engine assembly task from
a local company. The task is particularly interesting as one of the sub-tasks
is to insert a rocker shaft that weights 4.3 kg and would therefore
benefit from human-robot collaboration. The task is illustrated
in Figure~\ref{fig:assembly_steps} which also shows the five sub-tasks
(H denotes the human operator and R the robot):
Task~1) Install 8 rocker arms (H),
Task~2) Install the motor frame (R),
Task~3) Insert 4 frame screws (H),
Task~4) Install the rocker shaft (R+H) and
Task~5) Insert the nuts on the shaft (H).

Tasks~1-3 and 5 are dependent so that the previous sub-task must be completed before
the next can begin. Task~4 is collaborative in the sense that the robot brings
the saft and moves to a force mode allowing physical hand-guidance of the
end-effector. In the force mode, the robot applies just enough force to overcome
the gravitational force of the object while still allowing the human to guide the
robot arm for accurate positioning.     

\subsection{A non-collaborative baseline}
Our baseline is based on the current practices in manufacturing - the human
and robot cannot operate in the same workspace simultaneously. In our setting,
the operator must stay 4 meters apart from the robot when the robot is moving
and the operator is allowed to enter the workspace only when the robot is
not moving. In this scenario the collaborative Task~4 is completely
manual, the robot only brings the part.

Safety in the baseline is ensured by a \textit{dead man's switch} button which
the operator needs to press all the time for the robot to be operational.
The baseline does not contain any UI components, but in the user studies the
subjects are provided with textual descriptions of all sub-tasks.

\section{Experiments}
We report quantitative and qualitative results for the assembly task and
compare the three setups.

\subsection{Settings}
The experiments were conducted using the model 5 Universal Robot Arm (UR5) and
OnRobot RG2 gripper. KinectV2 was used as the depth sensor installed
to the ceiling and capturing the whole workspace area. The AR displays,
the projector or HoloLens, were connected to a single laptop with Ubuntu 16.04
OS and it performed all computations.

\begin{figure}[t]
  \begin{center}
    \includegraphics[width=0.9\linewidth]{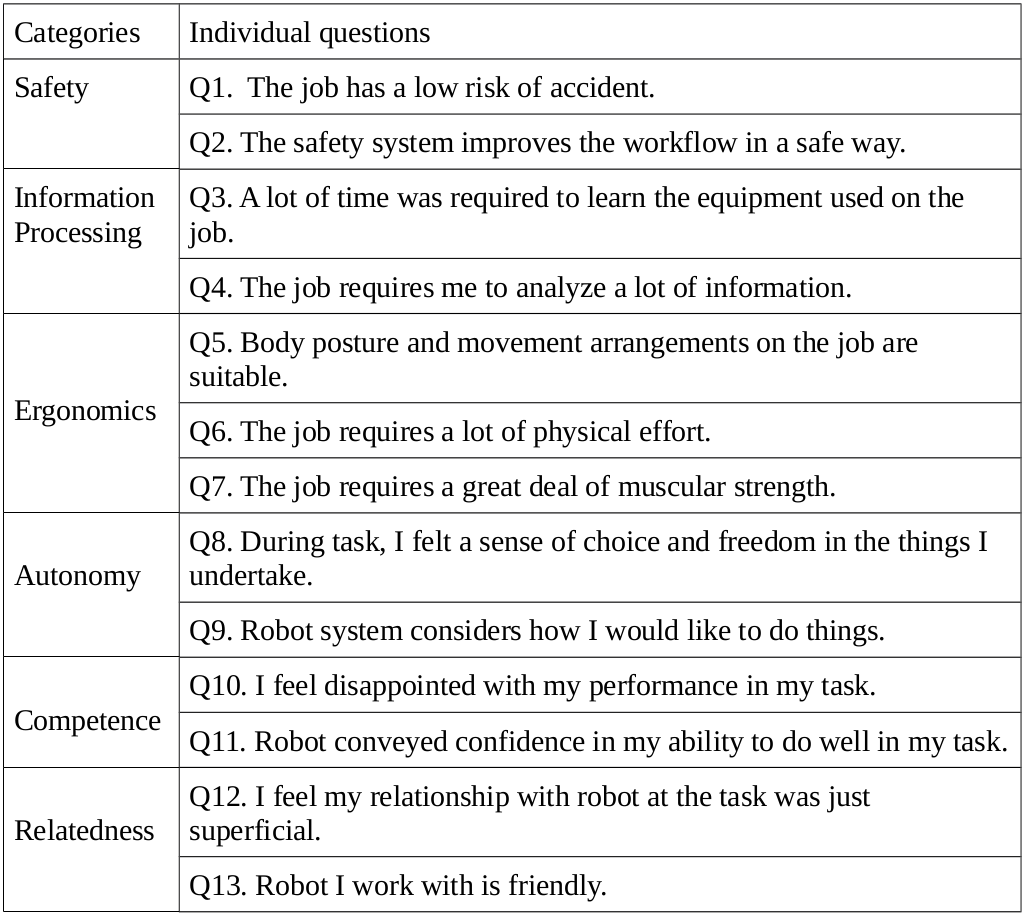}
  \end{center}

  \vspace{-1.5\medskipamount}
  
  \caption{The questionnaire used in our subjective experiments.}
  \label{fig:questionnaire}
\end{figure}

\paragraph{User studies}
The experiments were conducted with 20 unexperienced volunteered
university students. Their performance times were recorded and after experimenting
the three systems they were asked the questionnaire in
Figure~\ref{fig:questionnaire}. The goal of the questionnaire was to evaluate
physical and mental stress aspects of the human co-workers during the task.
The questions were selected to cover safety, ergonomics and mental stress
experience as defined in Salvendy et al.~\cite{salvendy2012handbook},
and autonomy, competence, and relatedness in Deci et al.~\cite{deci1991motivation}.
Users were asked to score each question using the scale from 1 (totally disagree)
to 5 (totally agree).
\subsection{Quantitative Performance}
\begin{figure}[h]
  \begin{center}
    \includegraphics[width=0.8\linewidth]{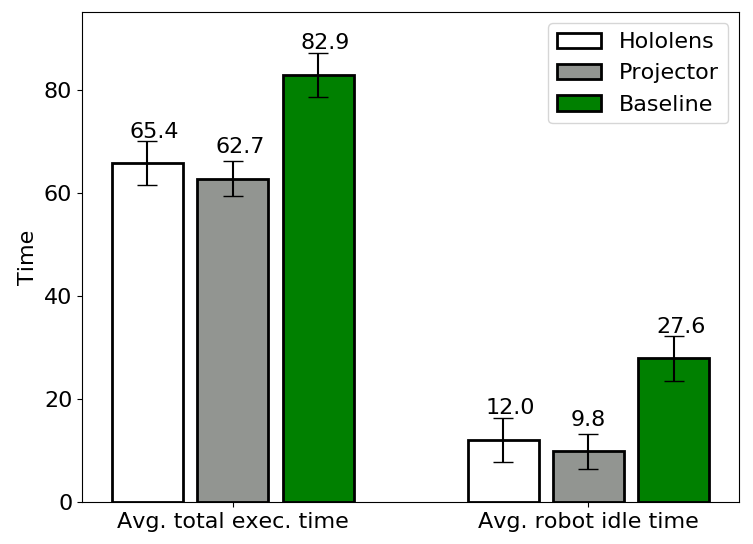}
    \caption{Average task execution and robot idle times from the user studies.}
    \label{fig:results1}
  \end{center}
\end{figure}
For quantitative performance evaluation we used two different
metrics, {\em Average total task execution time} and
{\em Average total robot idle time}, that measure the total
performance improvement and the time robot is waiting for
the operator to complete her tasks.

The results in Figure~\ref{fig:results1} show that the both AR-based
interactive systems outperform the baseline where the robot was not moving
in the same workspace with an operator. The difference can be explained
by the robot idle time which is much less for AR-based interaction. The
difference between the HoloLens and Projector based systems is marginal.
On average, the AR-based systems were $21-24\%$ and $57-64\%$ faster than the
baseline in the terms of the total execution time and the robot idle time respectively.

\begin{figure*}
	\includegraphics[width=1.0\linewidth]{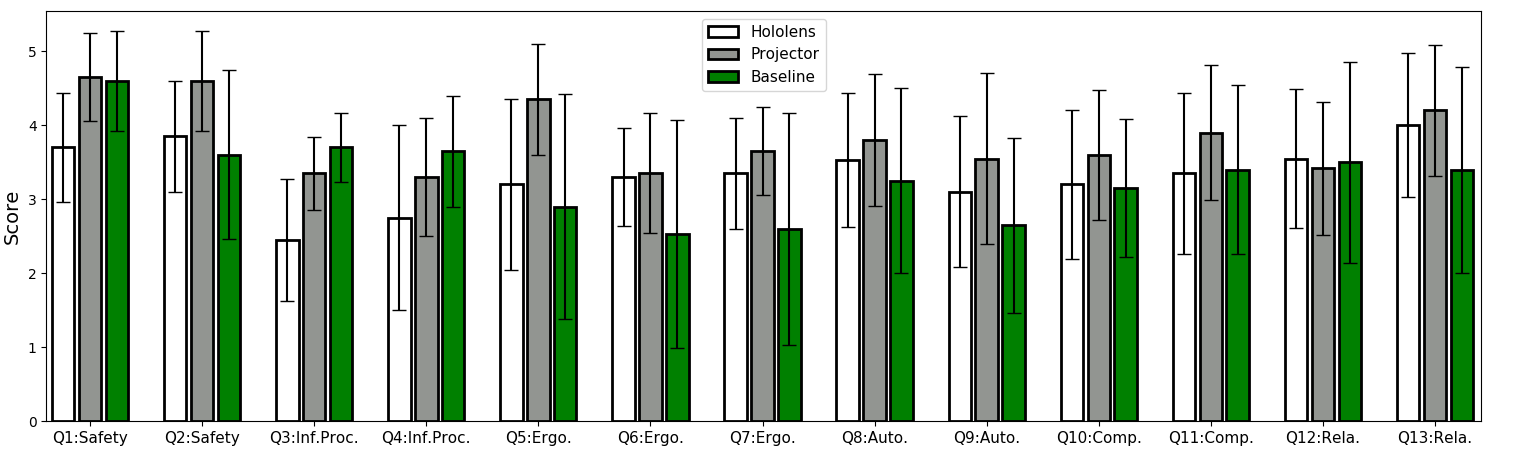}
	\caption{Average scores for the questions Q1-Q13 used in the user studies (20 participants).
          Score 5 denotes ``totally agree'' and 1 ``totally disagree'' and scores for the questions Q3, Q4, Q6, Q7
          and Q10 are inverted for better readability (score 5 has the same meaning as for other questions).
          \label{fig:results2}}
\end{figure*}

\subsection{Subjective evaluation}
Since the results from the previous quantitative evaluation of
system performance were similar for the both HoloLens and Projector
based AR interaction the user studies provided important information
about the differences of the two systems.

From the each 20 participant the 13 template
questions (Q1-Q13) in Figure~\ref{fig:questionnaire} were asked
and the results analyzed.
The average scores with the standard
deviations are shown in Figure~\ref{fig:results2}. The overall
impression is that the Projector-based display outperforms the two others
(HoloLens and Baseline), but surprisingly HoloLens is found inferior
the baseline in many safety related questions.

The numerical values are given in Table~\ref{tab:results2} and these verify the overall findings.
The projector-based method is considered the safest and the HoloLens-based method most unsafe with
a clear margin. The amount of information needed to understand the task is smallest for the baseline
while Projector-based has very similar numbers and again the HoloLens-based method was found clearly
more difficult to understand. Ergonomy-wise the HoloLens and Projector based methods were superior
likely to the fact that they provided help in installing the heavy rocker shaft.
The autonomy numbers are similar for all methods, but the projector-based is found the easiest to
work with. The users also found their performance best with the Projector-based system (Competence).
The question Q12 was obviously difficult to understand for the users, but all users found the
system with AR-interaction more plausible (Q13) than the baseline without interaction. Overall,
the projector-based AR interaction in collaborative manufacturing was found safer and more ergonomic
than the baseline without AR interaction and also the HoloLens-based AR.
\begin{table}[h]
  \caption{Average scores for the question (Q1-Q13). Higher is better except for those marked
  with ``$\neg$''. The best result emphasized (multiple if no statistical significance) \label{tab:results2}}
  \begin{center}
 \resizebox{0.9\linewidth}{!}{
   \begin{tabular}{llrrr}
    \toprule
    \multicolumn{1}{c}{}                                                                                   &                          & \multicolumn{1}{c}{{\em HoloLens}} & \multicolumn{1}{c}{{\em Projector}} & \multicolumn{1}{c}{{\em Baseline}} \\
    \midrule
    \multicolumn{1}{l}{\multirow{2}{*}{Safety}}                                                           & Q1                       & 3.7                          & {\bf 4.7}                           & {\bf 4.6}                          \\
    \multicolumn{1}{l}{}                                                                                  & Q2                       & 3.9                         & {\bf 4.6}                            & 3.6                          \\
    \midrule
    \multicolumn{1}{l}{\multirow{2}{*}{\begin{tabular}[c]{@{}l@{}}Information\\ Processing\end{tabular}}} & Q3$\neg$                       & 2.6                         &      1.7                             & {\bf 1.3}                          \\
    
    \multicolumn{1}{l}{}                                                                     & Q4$\neg$                       & 2.3                         &      1.7                             & {\bf 1.4}                          \\
    \midrule
    \multicolumn{1}{l}{\multirow{3}{*}{Ergonomics}}                                                       & Q5                       & 3.2            &      {\bf 4.4}                           & 2.9                          \\
    \multicolumn{1}{l}{}                                                                               & Q6$\neg$                         & {\bf 1.7}            &      {\bf 1.7}                           & 2.5                          \\
      \multicolumn{1}{l}{}                                                                                  & Q7$\neg$                       & 1.7               &      {\bf 1.4}                           & 2.4                          \\
      \midrule
      \multicolumn{1}{l}{\multirow{2}{*}{Autonomy}}                                                         & Q8                       & 3.5                   & {\bf 3.8}                           & 3.3                          \\
      \multicolumn{1}{l}{}                                                                                  & Q9                       & 3.1                   & {\bf 3.6}                           & 2.7                          \\
      \midrule
      \multicolumn{1}{l}{\multirow{2}{*}{Competence}}                                                       & Q10$\neg$                      & 1.8                   & {\bf 1.4}                           & 1.9                          \\
      \multicolumn{1}{l}{}                                                                                  & \multicolumn{1}{l}{Q11} & 3.4                   & {\bf 3.9}                           & 3.4                          \\
      \midrule
      \multicolumn{1}{l}{\multirow{2}{*}{Relatedness}}                                                      & \multicolumn{1}{l}{Q12} & {\bf 3.6}                          & {\bf 3.4}                           & {\bf 3.5}                          \\ 
	\multicolumn{1}{l}{}                                                                                  & \multicolumn{1}{l}{Q13} &    {\bf 4.0}                           & {\bf 4.2}                           & 3.4                          \\
        \bottomrule
  \end{tabular}}
  \end{center}
\end{table}

Below are free comments from the user studies that well point out the reasons why different
systems were preferred or considered difficult to use:
\begin{compactitem}
	\item \textbf{HoloLens:} ``\emph{Too narrow field of view, head has to be rotated a lot.}''\\ ``\emph{Feels heavy and uncomfortable after a while.}''\\ ``\emph{Holograms feels to be closer than they actually are.}''
	\item \textbf{Projector:} ``\emph{I would choose the projector system over HoloLens}''\\ ``\emph{Easier and more comfortable to use}''
	\item \textbf{Baseline:} ``\emph{System could be fooled by placing object on the switch button.}''
\end{compactitem}
%

\section{Conclusions}
We proposed a computation model of the shared workspace in human-robot collaborative
manufacturing. The model allows to monitor changes in the workspace to establish
safety features.
Moreover, we proposed a user interface for
HRC in industrial manufacturing and implemented it on two different hardware
for augmented reality, a project-mirror and wearable AR gear (HoloLens). In
our experiments on a realistic assembly task adopted from automotive manufacturing the both AR-based systems were found superior in performance
to the baseline without a shared workspace. However, the users found the projector-mirror
system clearly more plausible for manufacturing work than the HoloLens setup.

%
\bibliographystyle{ieeetr}
\bibliography{hrc_safety}

\begin{thebibliography}{10}

\bibitem{isots15066}
International Organization for Standardization, {\em ISO/TS 15066:2016 –
  Robots and Robotic Devices – Collaborative Robots}, 2016.

\bibitem{Marvel-2017-rcim}
J.~Marvel and R.~Norcross, ``Implementing speed and separation monitoring in
  collaborative robot workcells,'' {\em Robotics and Computer-Integrated
  Manufacturing}, vol.~44, 2017.

\bibitem{bdiwi2017new}
M.~Bdiwi, M.~Pfeifer, and A.~Sterzing, ``A new strategy for ensuring human
  safety during various levels of interaction with industrial robots,'' {\em
  {CIRP} Annals}, vol.~66, no.~1, 2017.

\bibitem{villani2018survey}
V.~Villani, F.~Pini, F.~Leali, and C.~Secchi, ``Survey on human--robot
  collaboration in industrial settings: Safety, intuitive interfaces and
  applications,'' {\em Mechatronics}, 2018.

\bibitem{Lasota-2017}
P.~Lasota, T.~Fong, and J.~Shah, ``A survey of methods for safe human-robot
  interaction,'' {\em Foundations and Trends in Robotics}, vol.~5, no.~4, 2017.

\bibitem{lasota2014toward}
P.~A. Lasota, G.~F. Rossano, and J.~A. Shah, ``Toward safe close-proximity
  human-robot interaction with standard industrial robots,'' 2014.

\bibitem{flacco2015depth}
F.~Flacco, T.~Kroeger, A.~De~Luca, and O.~Khatib, ``A depth space approach for
  evaluating distance to objects,'' {\em Journal of Intelligent \& Robotic
  Systems}, vol.~80, no.~1, pp.~7--22, 2015.

\bibitem{vogel2017safeguarding}
C.~Vogel, C.~Walter, and N.~Elkmann, ``Safeguarding and supporting future
  human-robot cooperative manufacturing processes by a projection-and
  camera-based technology,'' {\em Procedia Manufacturing}, vol.~11, pp.~39--46,
  2017.

\bibitem{icinco15}
G.~Dumonteil, G.~Manfredi, M.~Devy, A.~Confetti, and D.~Sidobre, ``Reactive
  planning on a collaborative robot for industrial applications,'' in {\em
  Proceedings of the 12th International Conference on Informatics in Control,
  Automation and Robotics - Volume 2: ICINCO,}, pp.~450--457, INSTICC,
  SciTePress, 2015.

\bibitem{unhelkar2018human}
V.~V. Unhelkar, P.~A. Lasota, Q.~Tyroller, R.-D. Buhai, L.~Marceau, B.~Deml,
  and J.~A. Shah, ``Human-aware robotic assistant for collaborative assembly:
  Integrating human motion prediction with planning in time,'' {\em IEEE
  Robotics and Automation Letters}, vol.~3, no.~3, pp.~2394--2401, 2018.

\bibitem{chadalavada2015s}
R.~T. Chadalavada, H.~Andreasson, R.~Krug, and A.~J. Lilienthal, ``That's on my
  mind! robot to human intention communication through on-board projection on
  shared floor space,'' in {\em Mobile Robots (ECMR), 2015 European Conference
  on}, pp.~1--6, IEEE, 2015.

\bibitem{andersen2016projecting}
R.~S. Andersen, O.~Madsen, T.~B. Moeslund, and H.~B. Amor, ``Projecting robot
  intentions into human environments,'' in {\em Robot and Human Interactive
  Communication (RO-MAN), 2016 25th IEEE International Symposium on},
  pp.~294--301, IEEE, 2016.

\bibitem{vogel2011towards}
C.~Vogel, M.~Poggendorf, C.~Walter, and N.~Elkmann, ``Towards safe physical
  human-robot collaboration: A projection-based safety system,'' in {\em
  IEEE/RSJ International Conference on Intelligent Robots and Systems (IROS)},
  pp.~3355--3360, IEEE, 2011.

\bibitem{vogel2013projection}
C.~Vogel, C.~Walter, and N.~Elkmann, ``A projection-based sensor system for
  safe physical human-robot collaboration,'' in {\em IEEE/RSJ International
  Conference on Intelligent Robots and Systems (IROS)}, pp.~5359--5364, IEEE,
  2013.

\bibitem{7942791}
D.~Q. Huy, I.~Vietcheslav, and G.~S.~G. Lee, ``See-through and spatial
  augmented reality - a novel framework for human-robot interaction,'' in {\em
  2017 3rd International Conference on Control, Automation and Robotics
  (ICCAR)}, pp.~719--726, April 2017.

\bibitem{Elsdon2018icra}
J.~Elsdon and D.~Y, ``Augmented reality for feedback in a shared control
  spraying task,'' in {\em {IEEE} International Conference on Robotics and
  Automation (ICRA)}, 2018.

\bibitem{park_robot_1994}
F.~C. Park and B.~J. Martin, ``Robot sensor calibration: solving {AX}= {XB} on
  the {Euclidean} group,'' {\em IEEE Transactions on Robotics and Automation},
  vol.~10, no.~5, pp.~717--721, 1994.

\bibitem{hawkins2013analytic}
K.~P. Hawkins, ``Analytic inverse kinematics for the universal robots
  ur-5/ur-10 arms,'' tech. rep., Georgia Institute of Technology, 2013.

\bibitem{graham1983finding}
R.~L. Graham and F.~F. Yao, ``Finding the convex hull of a simple polygon,''
  {\em Journal of Algorithms}, vol.~4, no.~4, pp.~324--331, 1983.

\bibitem{rusu2010semantic}
R.~B. Rusu, ``Semantic 3d object maps for everyday manipulation in human living
  environments,'' {\em KI-K{\"u}nstliche Intelligenz}, vol.~24, no.~4,
  pp.~345--348, 2010.

\bibitem{MarKamLen:2011}
I.~Martynov, J.-K. Kamarainen, and L.~Lensu, ``Projector calibration by
  {''}inverse camera calibration{''},'' in {\em Scandinavian Conference on
  Image Analysis (SCIA)}, 2011.

\bibitem{salvendy2012handbook}
G.~Salvendy, {\em Handbook of human factors and ergonomics}.
\newblock John Wiley \& Sons, 2012.

\bibitem{deci1991motivation}
E.~L. Deci, R.~J. Vallerand, L.~G. Pelletier, and R.~M. Ryan, ``Motivation and
  education: The self-determination perspective,'' {\em Educational
  psychologist}, vol.~26, no.~3-4, pp.~325--346, 1991.

\end{thebibliography}

\end{document}